\documentclass{article}

\usepackage[final]{timeseries_workshop}
\usepackage{float}
\usepackage[none]{hyphenat}
\usepackage[utf8]{inputenc} 
\usepackage[T1]{fontenc}    
\usepackage{hyperref}       
\usepackage{url}            
\usepackage{booktabs}       
\usepackage{amsfonts}       
\usepackage{nicefrac}       
\usepackage{microtype}      
\usepackage{xcolor}         
\usepackage{amsmath}
\usepackage{amssymb}
\usepackage{makecell}
\usepackage{graphicx,subcaption}
\usepackage{enumitem}

\setcitestyle{round}

\newcommand{\edit}[1]{{#1}}

\newcommand{\mb}[1]{\mathbf{#1}}
\newcommand{\real}[1]{\mathbb{R}^{#1}}

\newcommand{\ihet}{\texttt{IL-HET}}
\newcommand{\ihom}{\texttt{IL-HOM}}
\renewcommand{\b}[1]{\textbf{#1}}
\renewcommand{\it}[1]{\underline{\textit{#1}}}

\title{From RNNs to Foundation Models: An Empirical Study on Commercial Building Energy Consumption}

\author{%
    Shourya Bose\\
    UC Santa Cruz\\
    \texttt{shbose@ucsc.edu}\\
    \And
    Yijiang Li\\
    Argonne National Laboratory\\
    \texttt{yijiang.li@anl.gov}\\
    \And
    Amy Van Sant\\
    National Renewable Energy Laboratory\\
    \texttt{amy.vansant@nrel.gov}\\
    \And
    Yu Zhang\\
    UC Santa Cruz\\
    \texttt{zhangy@ucsc.edu}\\
    \And
    Kibaek Kim\\
    Argonne National Laboratory\\
    \texttt{kimk@anl.gov}
}

\begin{document}

\maketitle

\begin{abstract}

Accurate short-term energy consumption forecasting for commercial buildings is crucial for smart grid operations. While smart meters and deep learning models enable forecasting using past data from multiple buildings, data heterogeneity from diverse buildings can reduce model performance.
The impact of increasing dataset heterogeneity in time series forecasting, while keeping size and model constant, is understudied. We tackle this issue using the ComStock dataset, which provides synthetic energy consumption data for U.S. commercial buildings. Two curated subsets, identical in size and region but differing in building type diversity, are used to assess the performance of various time series forecasting models, including finetuned open-source foundation models (FMs).
The results show that dataset heterogeneity and model architecture have a greater impact on post-training forecasting performance than the parameter count. Moreover, despite the higher computational cost, finetuned FMs demonstrate competitive performance compared to base models trained from scratch.
\end{abstract}

\section{Introduction}
\label{sec:intro}
In smart grids, accurate forecasting of future load consumption across different timescales is critical for both operational tasks, such as generator scheduling, and long-term planning, such as capacity expansion. A key area of focus is short-term load forecasting (STLF), which predicts energy consumption over timeframes from an hour to a day, at the level of individual buildings~\citep{loadforecasting-review-ME-JAKS-RB-BDM:2007}. The canonical pipeline for STLF involves collecting smart meter data of multiple buildings into a single dataset, training a single model on the same, and validating it on data split along the time axis~\citep{lstmloadforecast-WK-etal:2019}. Pooling large datasets is particularly crucial for training large models, including foundation models, as it enables the models to learn diverse patterns and complex relationships from a wide variety of buildings. This global approach of training on pooled datasets requires forecasting models to effectively manage the heterogeneity that arises from differences in building data~\citep{globalmodels-HH-CB-KB:2022}.
The heterogeneity may manifest in scale, variance, seasonality, or trend of different time series.

We use the ComStock dataset~\citep{comstock} to study the effect of heterogeneity on various STLF models. 
It is a synthetic dataset containing energy consumption of the U.S. commercial building stock alongside additional features such as building type. We leverage these features to control heterogeneity in curated ComStock subsets by varying building type diversity. We generate two such datasets, with one exhibiting greater heterogeneity. In this context, we evaluate multiple time series forecasting models from the literature, broadly categorized into two types: base models and foundation models (FMs). Base models are trained from scratch on a given dataset, while FMs are pre-trained on a large corpora, and only need to be finetuned for the given downstream application. We sample across a wide cross-section of available base models: recurrent neural networks (RNNs) (LSTM~\citep{lstm-SH-JS:1997}, LSTNet~\citep{lstnet-GL-WCC-YY-HL:2018}), transformer-based architectures (Transformer~\citep{transformer-AV-etal:2017}, Informer~\citep{informer-HZ-SZ-JP-ZS-JL-HZ-WZ:2021}), decomposition-based architectures (Autoformer~\citep{autoformer-HW-JX-JW-ML:2021}), \edit{patch-based architectures (PatchTST~\citep{patchtst-TN-etal:2023})}, and 2D-backbone architectures (TimesNet~\citep{timesnet-HW-TH-YL-HZ-JW-ML:2023}). For foundation models, we select TimesFM~\citep{timesFM-AD-etal:2024}, which handles \edit{univariate} inputs and is tested in both zero-shot (ZS) and finetuned (FT) modes. This setup enables us to assess the benefits of pre-training alongside the impact of heterogeneity.

%
This paper makes three key contributions: (i) the introduction of two curated building energy consumption datasets from ComStock, which are of the same size but vary in heterogeneity; (ii) a performance comparison of various base models on these datasets; and (iii) an evaluation of foundation models in both zero-shot and finetuned modes, compared to base models.



\section{ComStock Dataset}\label{s:comstock}
\begin{figure}[tb!]
    \centering
    \hfill
    \begin{subfigure}[b]{0.3\linewidth}
        \includegraphics[height=3.65cm, keepaspectratio]{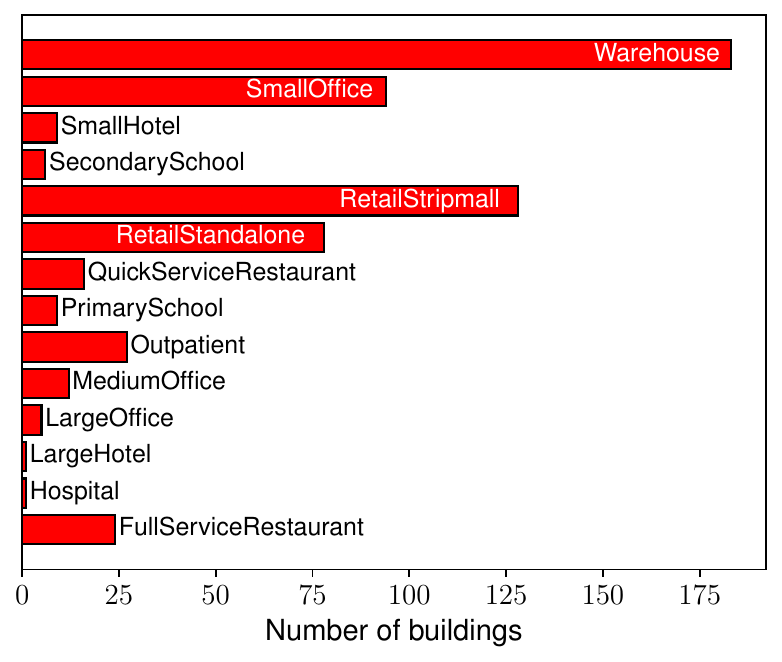}
        \caption{}
        \label{fig:barplot}
    \end{subfigure}
    \hfill
    \begin{subfigure}[b]{0.38\linewidth}
        \includegraphics[height=3.65cm, keepaspectratio]{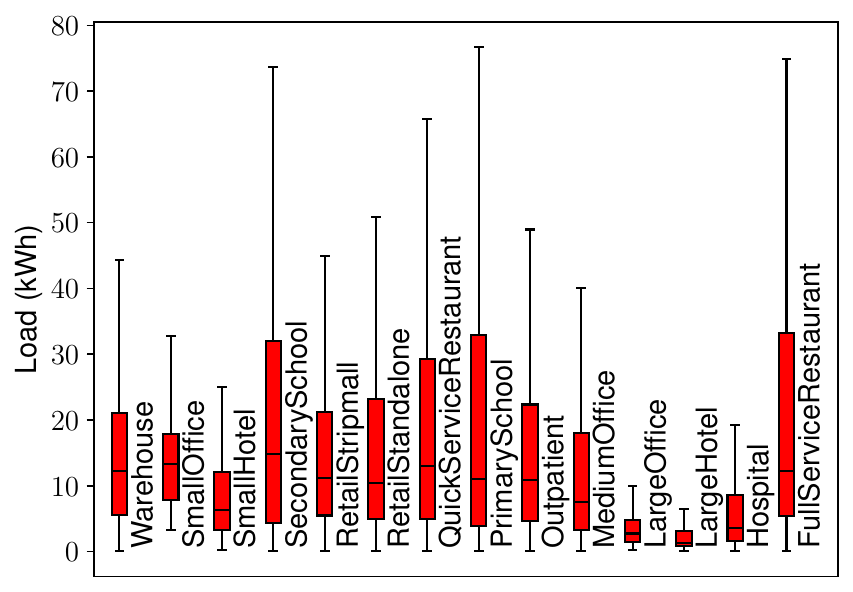}
        \caption{}
        \label{fig:spread}
    \end{subfigure}
    \hfill
    \begin{subfigure}[b]{0.3\linewidth}
        \includegraphics[height=3.65cm, keepaspectratio]{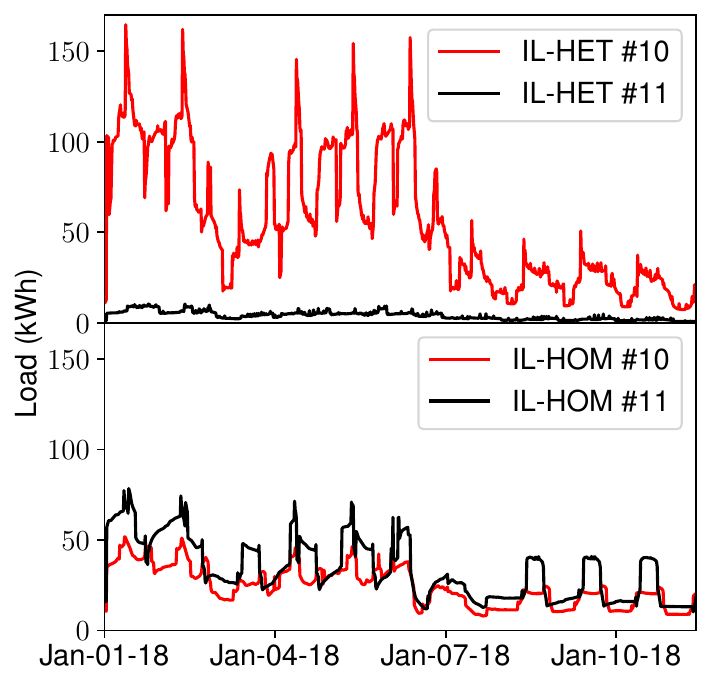}
        \caption{}
    \end{subfigure}
    \hfill
    \vspace{-0.2cm}
    \caption{(a) Distribution of building type in \ihet. (b) Boxplot of the spread of load values for each building type in \ihet. (c) Load shapes from two buildings each from \ihet\ and \ihom.}
    \vspace{-0.4cm}
\end{figure}

ComStock, developed by the National Renewable Energy Laboratory and funded by the U.S. Department of Energy, is a detailed model of the U.S. commercial building stock. It uses statistical sampling methods and advanced energy simulations via OpenStudio~\citep{openstudio} to estimate energy consumption across \edit{the building stock using} 350,000 unique building energy models. These models are created \edit{from over 80 input distributions, including} building characteristics \edit{such as} building type, construction year, HVAC system, and occupancy schedules. The dataset provides detailed outputs like energy consumption at \edit{15-minute resolution} across different end uses and critical metrics, resulting in approximately 10 TB of data per release.

For our study, we distill the ComStock data into two curated datasets from buildings in Illinois, termed \ihet\ (heterogeneous) and \ihom\ (homogeneous). Each dataset contains \edit{data from} 592 building \edit{models} for the actual meteorological year 2018. \ihet\ includes all 14 available building types in ComStock \edit{(see Figure~\ref{fig:barplot})}, maintaining the overall dataset distribution, thus introducing heterogeneity. In contrast, \ihom\ comprises only buildings of type \texttt{Warehouse}, providing a more homogeneous set due to correlations among similar building loads. The heterogeneity in \ihet\ is illustrated through its temporal spread in Figure~\ref{fig:spread}, while \ihom\ demonstrates a more uniform pattern (Appendix~\ref{sec:app-analysis-dataset}). \edit{Both datasets are available at at \url{https://huggingface.co/datasets/APPFL/Illinois_load_datasets}.}

Apart from the load in kWh, we incorporate several features in the dataset. We incorporate temporal information using two categorical features that represent the index of the 15-minute interval of the day and the day of the week. As mentioned earlier, we include the actual meteorological year 2018 weather features by the way of temperature and wind speed. Furthermore, we include three building-specific static features \emph{viz} floor space, wall area, and window area. This feature selection is broader than the one in conventional energy datasets (see Appendix~\ref{sec:app-analysis-dataset}), representative of the typical data used by utility companies~\citep{pge}, and is validated via correlation analysis on a separate selection of buildings not included in either \ihet\ or \ihom\ (see Appendix~\ref{sec:corr}).
\section{Experiment Design}
\label{sec:ed}
A pointwise time series forecasting problem can be generally represented as
\begin{align*}
    \hat{\mb{y}}_{t+T} = f_{\pmb{\theta}} (\mb{y}_{[t-L+1:t]},\mb{u}_{[t-L+1:t+T]},\mb{x}_{[t-L+1:t]},\mb{s}),
\end{align*}
where, for our experiments, $\mb{y}_t\in\real{1}$ is the building energy consumption, $\mb{x}\in\real{2}$ includes the two weather features, $\mb{u}_t\in\real{2}$ denotes two time induces, and $\mb{s}\in\real{3}$ includes the three static building characteristics, and $T$ and $L$ are the `lookahead' and `lookback' durations, respectively. The forecast model $f(\cdot)$ that is parameterized by the vector $\pmb{\theta}\in\real{h}$ can be trained in a supervised fashion, followed by evaluating its performance on unseen test sets. We use two metrics: \emph{\edit{normalized} mean squared error (NMSE)} and \emph{\edit{normalized} mean absolute error (NMAE)}, which given time indices of interest $t\in\mathcal{T}$, are defined as $
    \text{NMSE} := \frac{1}{n|\mathcal{T}|}\sum_{t\in\mathcal{T}} \left\lVert \hat{\mb{y}}_t - \mb{y}_t \right\rVert_2^2 / \sigma_{\mb{y}}^2$ and $\text{NMAE} := \frac{1}{n|\mathcal{T}|}\sum_{t\in\mathcal{T}} \left\lVert \hat{\mb{y}}_t - \mb{y}_t \right\rVert_1/\sigma_{\mb{y}}$. Here \edit{$\sigma_{\mb{y}}$ is the normalizing factor}. \edit{NMSE is used as the loss function for training}.

The dataset is divided into train, validation, and test sets in a ratio of $0.8:0.1:0.1$ along the time axis for each building. Each set is $z$-normalized \edit{feature-wise} using statistics from the train set, \edit{and the standard deviation of energy consumption is used as $\sigma_{\mb{y}}$}. Data from all buildings are then concatenated to form unified datasets. Our choice of models (cf. Section~\ref{sec:intro}) aims to provide a comprehensive evaluation of various time series models from the literature. Detailed descriptions of the model categories are provided in Appendix \ref{s:basearch}. \edit{Since the pretraining of TimesFM was done by~\cite{timesFM-AD-etal:2024} for a maximum lookback of 512, this becomes the limiting factor in choosing the lookback of all models. Correspondingly, all models use $L=512$. Furthermore, we only use the energy consumption feature with TimesFM since it is a univariate model.}
\edit{Owing to the diversity of base models as well as TimesFM, we carry out hyperparameter tuning for inferring the optimal learning rate for each model, heterogeneity level and lookahead separately, the details of which are provided in Appendix~\ref{s:tuning}. The base models are trained for 20 epochs, while TimesFM is finetuned using full-parameter fine tuning for 10 epochs. The impact of using parameter-efficient finetuning for TimesFM is in the scope of future work. All models are trained with a batch size of 1024, and the training of each model uses a node with 4 Nvidia A100 GPUs. The final trained model weights are available at \url{https://huggingface.co/APPFL/Building_load_forecasting}, while zero-shot experiments with TimesFM are performed using the off-the-shelf pre-trained \texttt{timesfm-1.0-200m} (\url{https://huggingface.co/google/timesfm-1.0-200m}) without any modification.}

\section{Experiment Results}
The performance of the trained models on the test set is summarized in Table~\ref{table:results}, and predictions on the test set are visualized in Appendix~\ref{sec:reconst}. \edit{Among the base models, the best performance on all experiments is achieved by TimesNet except one case where PatchTST outperforms it. This implies that the 2D-backbone of TimesNet and the periodicity-based patch generation of PatchTST are both effective in learning the patterns needed to generate high-quality forecasts. On the other hand, the relatively lower performance of Transformer, Autoformer, and Informer can be attributed to loss of temporal information in self-attention mechanisms and embeddings~\citep{are-transformers-effective}. This leads to such models being outperformed by simpler architectures such as LSTM and LSTNet. It is important to note that patch-based architectures such as PatchTST and TimesFM (whose underlying architecture is patch-based) performs better on \ihet\ than \ihom, which is a phenomenon that could indicate an increase in their performance with higher heterogeneity in the training data. This will be further investigated in future works. We also note that our implementation of early-stopping (i.e. terminating training when continued improvements are no longer observed on the test set) yielded disparate results across \ihet\ and \ihom\ on base models, with \ihet\ training terminating earlier across most models. This is discussed in Appendix~\ref{s:tuning}.}

\edit{Comparing the base models to TimesFM, we see that the fined-tuned TimesFM is able to outperform the best base models, and by a large margin on the \ihet\ dataset. It is also worth pointing out that TimesFM is a univariate model while the best base models, e.g., TimesNet and PatchTST, are multivariate models. This demonstrates the inherent benefit of the vast corpora of information absorbed by the model during pre-training when it comes to downstream forecasting tasks. Nonetheless, we should take the differences in model sizes between these models into account as well. On the other hand, the zero-shot performance of TimesFM is inadequate compared to other base models, highlighting the importance of full-parameter finetuning to unlock FM's potentials.}

\begin{figure}[tb!]
    \centering
    \hfill
    \begin{subfigure}[b]{0.28\linewidth}
        \includegraphics[height=3.3cm, keepaspectratio]{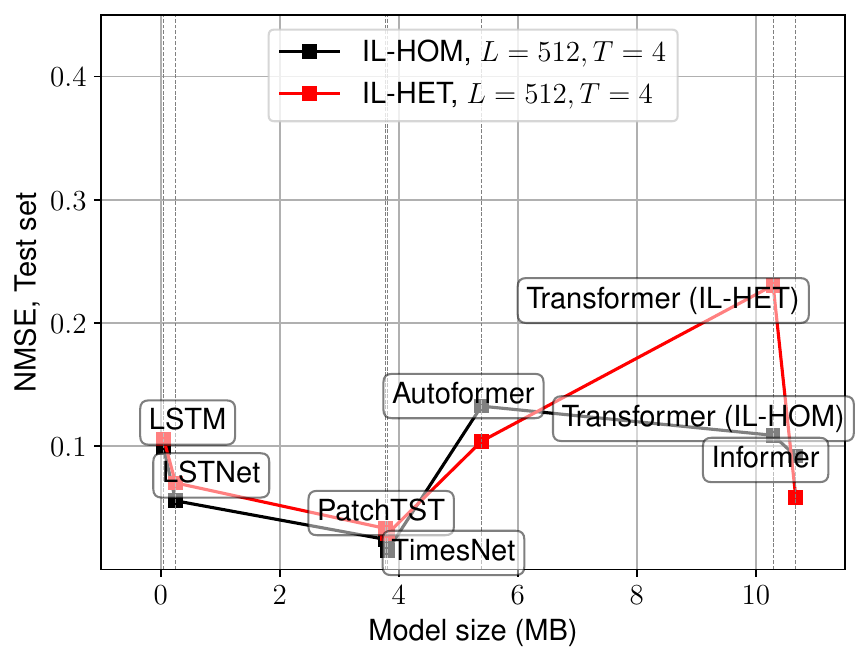}
        \caption{}
        \label{fig:four}
    \end{subfigure}
    \hfill
    \begin{subfigure}[b]{0.28\linewidth}
        \includegraphics[height=3.3cm, keepaspectratio]{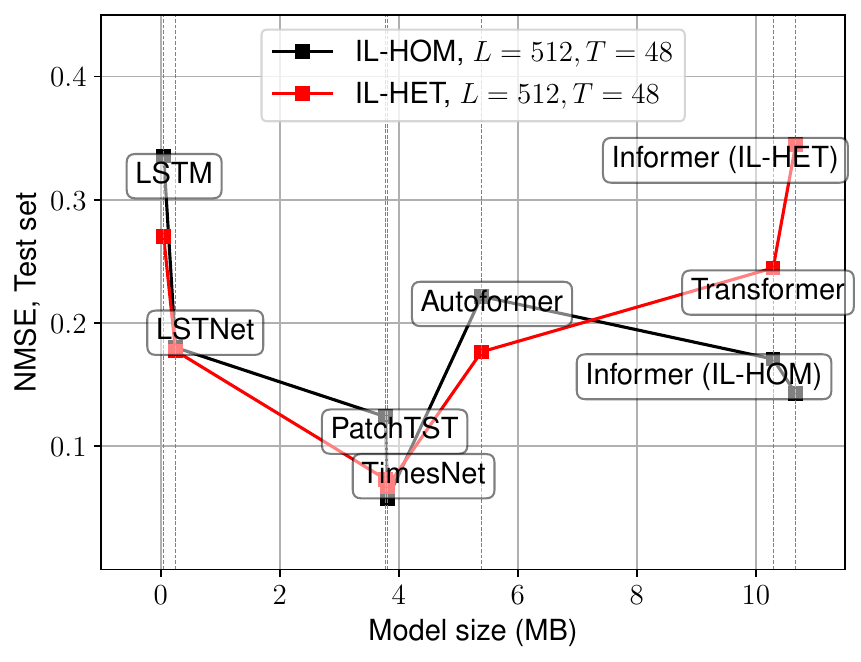}
        \caption{}
        \label{fig:fortyeight}
    \end{subfigure}
    \hfill
    \begin{subfigure}[b]{0.28\linewidth}
        \includegraphics[height=3.3cm, keepaspectratio]{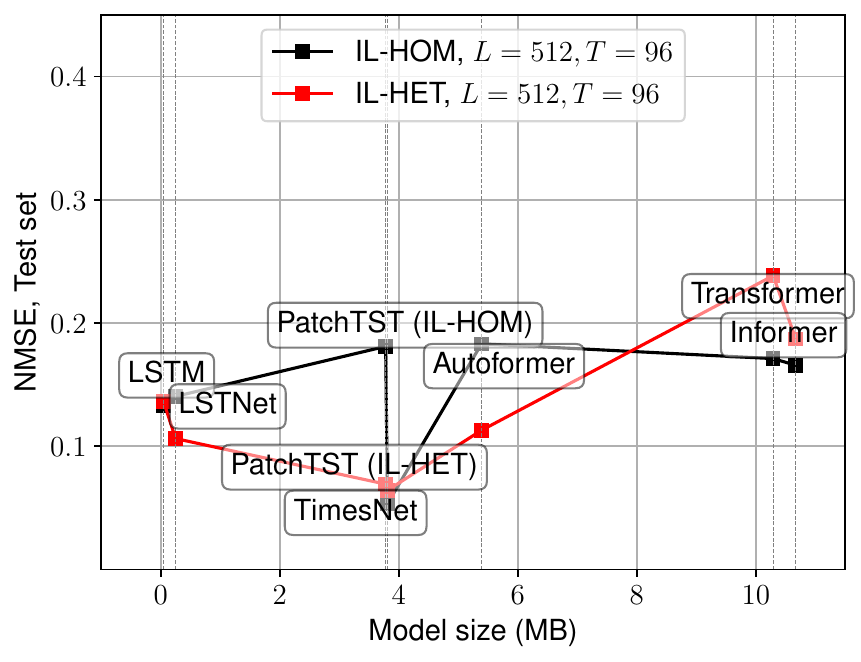}
        \caption{}
        \label{fig:ninetysix}
    \end{subfigure}
    \hfill
    \caption{Test set performance of base models for $L=512$ and (a) $T=4$, (b) $T=48$, (c) $T=96$.}
    \label{fig:size}
\end{figure}

Finally, we analyze the importance of base model size on performance in Figure~\ref{fig:size}. This not only visually reinforces the impact of increasing heterogeneity on model performance, but also highlights the point that model architecture is more important than brute-forcing model size. Evidently, the trend that relatively small models \edit{such as} LSTM, LSTNet, TimesNet, and \edit{PatchTST} can outperform bigger models highlights the importance of efficiently learning temporal relations and feature-dependence on forecasts, rather than increasing parameter sizes.
%

\begin{table}[h!]
  \caption{Experiment results for different models tested on the two datasets. The best results for each $(L,T)$ pair are in \b{bold}, while the second-best is \it{italicized and underlined}.}
  \label{table:results}
  \centering
  \resizebox{\textwidth}{!}{
  \begin{tabular}{c|c|c|c|c|c|c|c|c|c|c}
    \toprule
    \multicolumn{11}{c}{\ihet}\\
    \midrule
    $(L,T)$ & \multicolumn{2}{c}{\b{LSTM}} & \multicolumn{2}{c}{\b{LSTNet}} & \multicolumn{2}{c}{\b{Transformer}} & \multicolumn{2}{c}{\b{Autoformer}} & \multicolumn{2}{c}{\b{Informer}} \\
    & NMSE & NMAE & NMSE & NMAE & NMSE & NMAE & NMSE & NMAE & NMSE & NMAE\\
    \midrule
    $(\edit{512},4)$& \edit{0.1059} & \edit{0.1623} & \edit{0.0702} & \edit{0.1280} & \edit{0.2306} & \edit{0.2316} & \edit{0.1040} & \edit{0.1689} & \edit{0.5850} & \edit{0.6808} \\
    $(\edit{512},48)$& \edit{0.3464} & \edit{0.2701} & \edit{0.1512} & \edit{0.1774} & \edit{0.2551} & \edit{0.2447} & \edit{0.1208} & \edit{0.1766} & \edit{0.3248} & \edit{0.3451} \\  
    $(\edit{512},96)$& \edit{0.1361} & \edit{0.1568} & \edit{0.1062} & \edit{0.1449} & \edit{0.2384} & \edit{0.2299} & \edit{0.1129} & \edit{0.1727} & \edit{0.1871} & \edit{0.1808} \\    
    \cmidrule{0-8}
    & \multicolumn{2}{c}{\b{TimesNet}} & \multicolumn{2}{c}{\b{PatchTST}} & \multicolumn{2}{c}{\b{TimesFM (ZS)}} & \multicolumn{2}{c}{\b{TimesFM (FT)}}\\
    & NMSE & NMAE & NMSE & NMAE & NMSE & NMAE & NMSE & \multicolumn{1}{c}{NMAE} & \multicolumn{2}{c}{} \\
    \cmidrule{0-8}
    $(\edit{512},4)$& \edit{\it{0.0289}} & \edit{\it{0.0588}} & \edit{0.0332} & \edit{0.0598} & \edit{0.0371} & \edit{0.0598} & \edit{\b{0.0078}} & \multicolumn{1}{c}{\edit{\b{0.0295}}} \\
    $(\edit{512},48)$& \edit{\it{0.0674}} & \edit{\it{0.0924}} & \edit{0.0732} & \edit{0.0953} & \edit{0.1072} & \edit{0.1159} & \edit{\b{0.0341}} & \multicolumn{1}{c}{\edit{\b{0.0687}}} \\
    $(\edit{512},96)$& \edit{0.0642} & \edit{0.0892} & \edit{\it{0.0641}} & \edit{\it{0.0870}} & \edit{0.1050} & \edit{0.1125} & \edit{\b{0.0339}} & \multicolumn{1}{c}{\edit{\b{0.0657}}}\\
    \cmidrule{0-8}
    \midrule
    \multicolumn{11}{c}{\ihom}\\
    \midrule
    $(L,T)$ & \multicolumn{2}{c}{\b{LSTM}} & \multicolumn{2}{c}{\b{LSTNet}} & \multicolumn{2}{c}{\b{Transformer}} & \multicolumn{2}{c}{\b{Autoformer}} & \multicolumn{2}{c}{\b{Informer}} \\
    & NMSE & NMAE & NMSE & NMAE & NMSE & NMAE & NMSE & NMAE & NMSE & NMAE\\
    \midrule
    $(\edit{512},4)$& \edit{0.0945} & \edit{0.1457} & \edit{0.0557} & \edit{0.1259} & \edit{0.1058} & \edit{0.1572} & \edit{0.1325} & \edit{0.1994} & \edit{0.0920} & \edit{0.1538} \\
    $(\edit{512},48)$& \edit{0.3355} & \edit{0.2858} & \edit{0.1891} & \edit{0.2148} & \edit{0.1707} & \edit{0.2081} & \edit{0.2216} & \edit{0.2762} & \edit{0.1428} & \edit{0.1966} \\  
    $(\edit{512},96)$& \edit{0.1326} & \edit{0.1868} & \edit{0.1404} & \edit{0.1997} & \edit{0.1711} & \edit{0.2110} & \edit{0.1831} & \edit{0.2358} & \edit{0.1657} & \edit{0.2264}  \\    
    \cmidrule{0-8}
    & \multicolumn{2}{c}{\b{TimesNet}} & \multicolumn{2}{c}{\b{PatchTST}} & \multicolumn{2}{c}{\b{TimesFM (ZS)}} & \multicolumn{2}{c}{\b{TimesFM (FT)}}\\
    & NMSE & NMAE & NMSE & NMAE & NMSE & NMAE & NMSE & \multicolumn{1}{c}{NMAE} & \multicolumn{2}{c}{} \\
    \cmidrule{0-8}
    $(\edit{512},4)$& \edit{\it{0.0182}} & \edit{\it{0.0684}} & \edit{0.0432} & \edit{0.0868} & \edit{0.0559} & \edit{0.1025} & \edit{\b{0.0054}} & \multicolumn{1}{c}{\edit{\b{0.0309}}}\\
    $(\edit{512},48)$& \edit{\it{0.0572}} & \edit{\it{0.1230}} & \edit{0.1239} & \edit{0.1864} & \edit{0.3111} &\edit{0.2873} & \edit{\b{0.0417}} & \multicolumn{1}{c}{\edit{\b{0.1003}}}\\
    $(\edit{512},96)$& \edit{\it{0.0532}} & \edit{\it{0.1163}} & \edit{0.1166} & \edit{0.1810} & \edit{0.2911} & \edit{0.2686} & \edit{\b{0.0495}} & \multicolumn{1}{c}{\edit{\b{0.1065}}}\\
    \cmidrule{0-8}
  \end{tabular}
  }
\end{table}
\section{Concluding Remarks}
%
%
%
In this paper, we tested various base models, as well as a finetuned and zero-shot FM on two datasets, one more heterogeneous than the other. The results demonstrate the impact of heterogeneity on model performance, the capability of models like TimesNet \edit{and PatchTST} to be robust against heterogeneity, and the superior capability of finetuned FMs on both datasets. These observations serve towards grounding the problem of heterogeneity in time series models and act as stepping stones in domains such as federated learning, where data heterogeneity is inherent to the domain.

\acksection
This material is based upon work supported by the U.S. Department of Energy, Office of Science, under contract number DE-AC02-06CH11357. This research used resources of the Argonne Leadership Computing Facility at Argonne National Laboratory, which is supported by the Office of Science of the U.S. Department of Energy under contract DE-AC02-06CH11357. We also gratefully acknowledge the computing resources provided on Swing, a high-performance computing cluster operated by the Laboratory Computing Resource Center at Argonne National Laboratory.

\bibliographystyle{plainnat}
\bibliography{./refs.bib}

\newpage
\appendix
\section{Analysis of \ihet\ and \ihom}
\label{sec:app-analysis-dataset}
\begin{table}[H]
    \centering
    \caption{Statistics for loads in \ihom\ and \ihet.}
    \label{tab:my_label}    
    \begin{tabular}{ccc}
        \toprule
        \b{Dataset} & \b{Mean (loads)} & \b{Standard deviation (loads)}  \\
        \midrule
        \ihom & 17.4307 & 24.3129 \\
        \ihet & 21.8440 & 51.3635 \\
        \bottomrule
    \end{tabular}
\end{table}
Both \ihet\ and \ihom\ contain load values of 592 buildings, along with 7 other features. Since the data is recorded at a granularity of $15$ minutes for one year, the total number of data points are 35060. Thus, each of the datasets contain a total of 166,044,160 observations. Considering the fact that both datasets contain the same number of points, the higher heterogeneity of \ihet\ can be seen by compaing the means and standard deviations of the loads.\par
\begin{figure}[H]
    \centering
    \begin{subfigure}[b]{0.35\linewidth}
        \includegraphics[height=3.65cm, keepaspectratio]{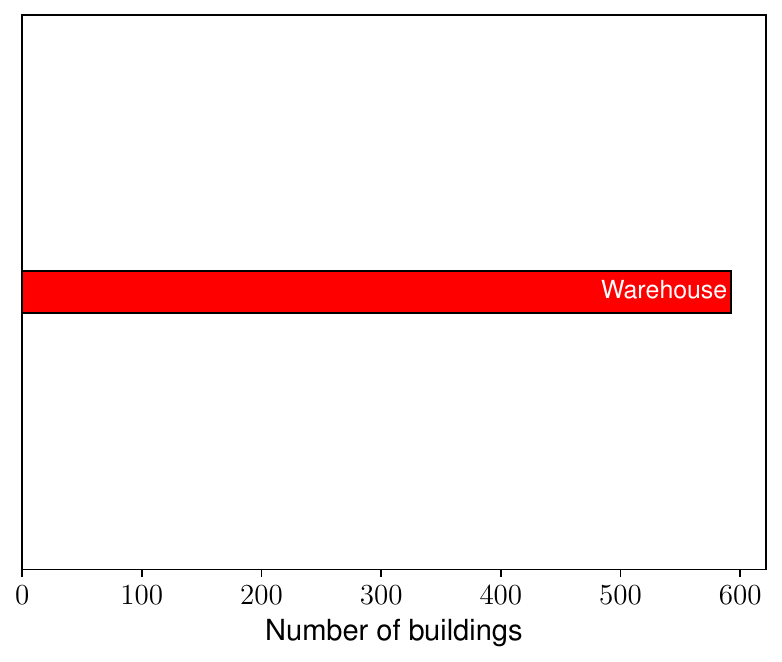}
        \caption{}
        \label{fig:barplottwo}
    \end{subfigure}
    \begin{subfigure}[b]{0.35\linewidth}
        \includegraphics[height=3.65cm, keepaspectratio]{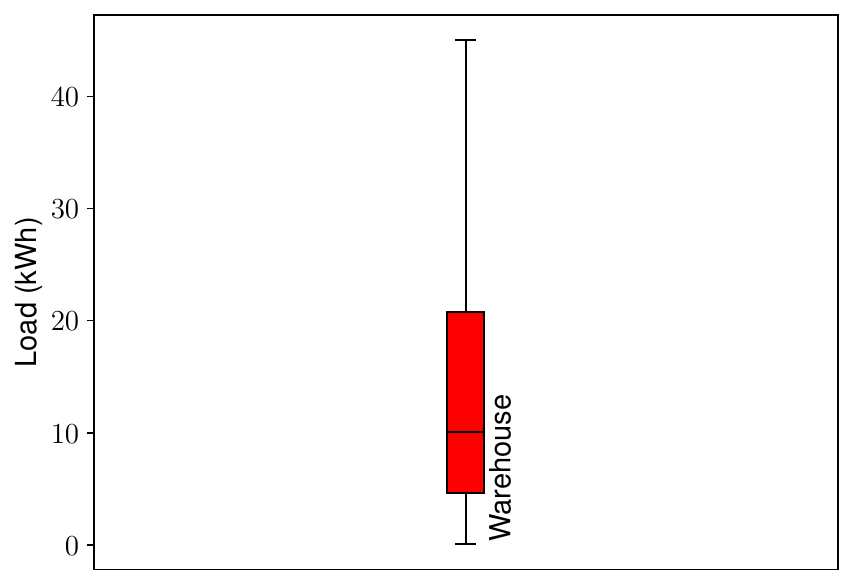}
        \caption{}
        \label{fig:spreadtwo}
    \end{subfigure}
    \hfill
    \caption{(a) Distribution of building type in \ihom. (b) Boxplot of the spread of values in the time series of each building type in \ihom.}
\end{figure}
Further evidence of the homogeneity of \ihom\ can be inferred from the equivalent figures to Figures~\ref{fig:barplot} and~\ref{fig:spread} as presented in Figures~\ref{fig:barplottwo} and~\ref{fig:spreadtwo}. While both figures contain only one element thanks to \ihom\ only containing buildings of the type \texttt{Warehouse}, the lower spread as well as variance of loads is evident from Figure~\ref{fig:spreadtwo} as compared to \ihet.
\begin{table}[H]
    \centering
    \caption{\ihom\ and \ihet\ compared to other building energy datasets.}
    \label{tab:competition}
    \begin{tabular}{cccccc}
     \toprule
     \textbf{Dataset} & \textbf{\# buildings} & \makecell{\textbf{Sampling}\\\textbf{frequency}} & \textbf{Duration} & \makecell{\textbf{Total}\\\textbf{load}\\\textbf{observations}} & \makecell{\textbf{Non-time}\\\textbf{features}}\\
     \midrule
      \ihom  & 592 & 15 minutes & 1 year & 20,755,520 & 5 \\
      \ihet & 592 & 15 minutes & 1 year & 20,755,520 & 5 \\
      \makecell{Electricity\\Load Diagrams} & 370 & 15 minutes & 4 years & 51,894,720 & 0\\
      \makecell{London\\Smart Meters} & 5,567 & 30 minutes & $\sim$2 years & 168,000,000 & 0\\
      \makecell{Buildings-900K} & 900,000 & 1 hour & 2 years & 15,777,000,000 & 3\\
    \bottomrule
    \end{tabular}
\end{table}
We compare \ihom\ and \ihet\ to other building-level load datasets, including Electricity Load Diagrams~\citep{electricityloaddiagrams}, Low Carbon London~\citep{lowcarbonlondon}, and Buildings-900K~\citep{buildingsbench}. While other datasets exceed ours in terms of the number of data points, we provide a larger number of features as shown in Table~\ref{tab:competition}, which provides important context to load forecasting models. As a matter of fact, the only other dataset on the list with non-time features is Buildings-900K which includes the latitude/longitude of the building's PUMA (Public Use Microdata Area) and a binary indicator for whether the building is commercial or residential.

\section{Correlation Analysis for Feature Selection}
\label{sec:corr}
In this section, we discuss our choice of features shown in Section~\ref{s:comstock}. For this analysis, we consider 42 randomly selected Illinois buildings from ComStock which are not in either \ihom\ or \ihet\ (which acts as a kind of `validation set' aiding in the construction of the datasets). We split our analysis into two parts: choice of weather (dynamic) features and choice of static features. Weather data available with ComStock includes seven features \emph{viz.} four temperature/wind features and three heat radiation features. We ignore the latter since they are not available in typical meteorological reports. Among the former four features, we calculate their Pearson correlation coefficient with the load values, and average the correlation across all 42 buildings. The results are presented in Table~\ref{tab:weather}.\par
\begin{table}[H]
    \centering
    \caption{Correlation analysis of weather features.}
    \label{tab:weather}
    \begin{tabular}{lc}
         \toprule
         \textbf{Feature name} & \textbf{Average correlation coefficient}\\
         \midrule
         Dry Bulb Temperature [${}^\circ C$] & 0.2507 \\
         Wind Speed [$m/s$] & 0.1396\\
         Wind Direction [Deg] & 0.0787 \\
         Relative Humidity [\%] & -0.2794\\
         \bottomrule
    \end{tabular}
\end{table}
From Table~\ref{tab:weather}, we select Dry Bulb Temperature [${}^\circ C$] and Wind Speed [$m/s$], since Wind Direction [Deg] is almost uncorrelated with load values and Relative Humidity [\%] is negatively correlated. On the other hand, ComStock contains many static features to choose from. In order to calculate static features' correlation to loads, we average the loads across all time indices and calculate the correlation of the averaged loads with the building's static features.\par
\begin{table}[H]
    \centering
    \caption{Correlation analysis of static features.}
    \label{tab:static}
    \begin{tabular}{lc}
    \toprule
         \textbf{Feature name} & \textbf{Correlation coefficient} \\
         \midrule
         External window area ($m^2$) & 0.9714\\
         Floor space ($\text{ft}^2$) & 0.9245\\
         External wall area ($m^2$) & 0.7385 \\
         Number of air loops & 0.7123 \\
         Total natural gas consumption & 0.7112\\
         Hot water supply volume ($m^3$) & 0.2672\\
         Occupant equivalent full load hour (EFLH) & 0.2672\\
         HVAC cooling type & 0.1386\\
         \bottomrule
    \end{tabular}
\end{table}
We choose the top three correlated features in order to prevent overcrowding of input data to the models with static features. Thus, the aforementioned analysis informs our choice of features, which in turn determines the models' learning capabilities.

\section{Model Architectures}\label{s:basearch}
\begin{table}[b!]
  \caption{Model sizes, parameter counts and structural hyperparameters for ($T=\edit{512},L=48$).}
  \label{table:features}
  \centering
  \begin{tabular}{cccc}
    \toprule
    \textbf{Model name} & \textbf{Model hyperparameters} & \makecell{\textbf{Parameter count}} & \textbf{Model size}\\
    \midrule
    \textbf{LSTM} & \makecell{Hidden size: $32$\\LSTM Layers: 1\\Dropout: 0.1} & 13,856 & 0.05 MB\\
    \midrule
    \textbf{LSTNet} & \makecell{Conv. layer output channels: 32\\Conv. kernel size: 12  \\GRU hidden size: 128\\Recurrent layer skips: $\{4,4\}$\\Attention window size: 7\\Dropout: 0.1} & 66,473 & 0.25 MB\\
    \midrule
    \textbf{Transformer} & \makecell{Token size: 128\\Num. heads: 4\\Encoder layers: 3\\Decoder layers: 3\\FCNN hidden dimension: 512\\Activation: GeLU\\Dropout: 0.1} & 2,698,625 & 10.29 MB\\
    \midrule
    \textbf{Autoformer} & \makecell{Token size: 128\\Num. heads: 4\\Encoder layers: 3\\Decoder layers: 3\\FCNN hidden dimension: 512\\Moving average window: 25\\TA Delay top-$k$: $5\log(L)$ \\Dropout: 0.1} & 1,413,633 & 5.39 MB\\
    \midrule
    \textbf{Informer} & \makecell{Token size: 128\\Num. heads: 4\\Encoder layers: 3\\Decoder layers: 3\\FCNN hidden dimension: 512\\Activation: GeLU\\Dropout: 0.1} & 2,798,211 & 10.67 MB\\
    \midrule
    \textbf{TimesNet} & \makecell{Token size: 128\\Encoder layers: 3\\$k$ for top-$k$ amplitudes: 5\\Num. kernels: 3\\Inception block intermediate channels: 64\\Dropout: 0.1} & 998,289 & 3.81 MB\\
    \midrule
    \edit{\b{PatchTST}} & \edit{\makecell{Token size: 128\\Num. heads: 3\\Encoder layers: 3\\Patch length: 16\\Stride: 8\\FCNN hidden dimension: 512}} & \edit{990,384} & \edit{3.78MB}\\
    \bottomrule
  \end{tabular}
\end{table}

Our choice of base models is meant to capture the major trends towards time series forecasting using deep architectures in recent years. To that end, we summarise these trends and the positioning of our chosen architectures within those trends.
\begin{itemize}[leftmargin=*]
    \item \textbf{Recurrent Neural Network Architectures:} These models consist of a learnable kernel that is unrolled recursively in time to generate forecasts. In this category, we use long-short term memory (LSTM)~\citep{lstm-SH-JS:1997} with a dense decoder, and LSTNet~\citep{lstnet-GL-WCC-YY-HL:2018}, which combines RNNs with convolutional layers.
    \item \textbf{Transformer-based Architectures:} These models use self-attention and cross-attention mechanisms in an encoder-decoder framework to parallelize the generation of the output sequence, given an input sequence. In this category, we use the vanilla Transformer~\citep{transformer-AV-etal:2017} and Informer~\citep{informer-HZ-SZ-JP-ZS-JL-HZ-WZ:2021}, with the latter incorporating the \emph{ProbSparse} attention mechanism. Contrary to conventional implementations, we use positional embeddings for temporal inputs (i.e. $\mb{u}$), alongside the standard positional embeddings.
    \item \textbf{Decomposition-based Architectures:} These models replace the attention mechanism of the Transformer with other mechanisms that leverage inherent characteristics of sequential data. In this category, we use Autoformer~\citep{autoformer-HW-JX-JW-ML:2021}, which uses an auto-correlation mechanism, alongside the decomposition of sequences into seasonal and trend patterns.
    \item \edit{\textbf{2D-backbone architectures:} These models convert input 1D temporal sequences into 2D tensors with properties that help the model better identify short- as well as long-term patterns in said sequences. In this category, we use TimesNet~\citep{timesnet-HW-TH-YL-HZ-JW-ML:2023}, which uses the Fourier transform to identify periodicity, which is then leveraged for tensor conversion. Following the same, the tensors are passed through 2D convolution-based Inception blocks~\citep{inception-CS-etal:2015} to generate the forecast.}
    \item \edit{\textbf{Patch-based Architectures:} These models split the input sequence into smaller sub-sequences called \emph{patches}, followed by learning the interactions between different patches to better predict the forecast. In this category, we use PatchTST~\citep{patchtst-TN-etal:2023}, which uses patching in addition to channel-independence to generate forecasts.}
\end{itemize}

A detailed description of the model sizes, parameter counts, and structural hyperparameters for the aforementioned models is provided in Table~\ref{table:features}. 

\edit{On the other hand,} TimesFM is a decoder-only foundation model built for time series forecasting. The model consists of input layers, stacked transformers, and output layers and considers an NMSE training loss function. The pre-trained TimesFM model \texttt{timesfm-1.0-200m} is trained on a vast amount and variety of data with both synthetic data and real-world data. There are two major model hyperparameters, namely, \texttt{model\_dim}, which determines the size of processed data, and \texttt{num\_layers}, which controls the number of stacked transformers in the model. A summary of the model size, parameter counts, and structural hyperparameters in TimesFM is provided in Table~\ref{table:timesFM_features}.

\begin{table}[H]
  \caption{Model sizes, parameter counts and structural hyperparameters for TimesFM (for a comprehensive list, see~\citep{timesFM-AD-etal:2024})}
  \label{table:timesFM_features}
  \centering
  \begin{tabular}{cccc}
    \toprule
    \textbf{Model name} & \textbf{Model hyperparameters} & \makecell{\textbf{Parameter count}} & \textbf{Model size}\\
    \midrule
    \textbf{TimesFM} & \makecell{Token size: $1280$\\ Num. Layers: $20$} & 200m & 854MB\\
    \bottomrule
  \end{tabular}
\end{table}

\section{Hyperparameter Tuning}\label{s:tuning}
Given these models span a wide range of architectures, using the same value for the learning rate for all of them would lead to poor performance. To that end, we tuned the learning rate over the validation set \edit{for all lookaheads ($T=4,48,96$) and all datasets (\ihet\ and \ihom)} with choices \edit{$\texttt{lr}=\{10^{-3},5\times10^{-4},10^{-4},5\times10^{-5},10^{-5},5\times10^{-6},10^{-6},5\times10^{-7}\}$}. The results are reported in Tables~\ref{tab:firstHyp} and~\ref{tab:secondHyp}.\par
\begin{table}[H]
    \centering
    \caption{Learning rate tuning over the validation set on the dataset \ihet. The metric is NMSE, and the best learning rates are highlighted.}
    \label{tab:firstHyp}
    \resizebox{\textwidth}{!}{
    \edit{
    \begin{tabular}{ccccccccc}
    \toprule
    \b{Model} & $10^{-3}$ & $5\times 10^{-4}$ & $10^{-4}$ & $5\times 10^{-5}$ & $10^{-5}$ & $5\times10^{-6}$ & $10^{-6}$ & $5\times 10^{-7}$\\
    \midrule
    \multicolumn{9}{c}{$L=512,T=4$}\\
    \midrule
    \b{LSTM} & 0.8258 & 0.2535 & 0.2020 & \b{0.1853} & 0.2804 & 0.2823 & 0.3878 & 0.4077 \\
    \b{LSTNet} & 0.4564 & 0.2236 & \b{0.1417} & 0.1709 & 0.2202 & 0.3225 & 0.3207 & 0.3397\\
    \b{Transformer} & 3.0464 & 3.0721 & 0.8025 & 1.2834 & 1.5921 & 1.0574 & \b{0.4246} & 0.6596\\
    \b{Autoformer} & 0.6904 & 0.5245 & 0.5835 & 0.4326 & 0.3083 & 0.3834 & 0.2688 & \b{0.2649}\\
    \b{Informer} & 3.9926 & 3.7363 & 3.0798  & 3.1364 & 1.5351 & \b{0.7345} & 1.0424 & 1.4432\\
    \b{TimesNet} & 0.0699 & 0.0667 & 0.0728 & 0.0627 & \b{0.0603} & 0.0703 & 0.1000 & 0.1088\\
    \b{PatchTST} & 0.0678 & 0.0622 & 0.0628 & \b{0.0597} & 0.0619 & 0.0666 & 0.0829 & 0.0924\\
    \midrule
    \multicolumn{9}{c}{$L=512,T=48$}\\
    \midrule
    \b{LSTM} & 3.8491 & 2.1324 & 0.3300 & \b{0.2981} & 0.3283 & 0.3052 & 0.3924 & 0.4104\\
    \b{LSTNet} & 0.2866 & 0.3833 & 0.2166 & \b{0.2125} & 0.3213 & 0.3544 & 0.3315 & 0.3432\\
    \b{Transformer} & 3.3272 & 3.3638 & 0.7665 & 0.8136 & 1.6502 & 1.0295 & \b{0.4785} & 0.7056\\
    \b{Autoformer} & 0.6104 & 0.5294 & 0.5150 & 0.4226 & 0.4226 & 0.3457 & \b{0.2744} & 0.2761\\
    \b{Informer} & 2.8727 & 3.0340 & 3.3786 & 3.6050 & 1.2244 & \b{0.3451} & 1.4153 & 1.9887\\
    \b{TimesNet} & 0.1043 & 0.1096 & 0.1117 & 0.1032 & \b{0.0976} & 0.0985 & 0.1121 & 0.1175\\
    \b{PatchTST} & 0.0967 & \b{0.0953} & 0.1009 & 0.0955 & 0.0984 & 0.1034 & 0.1082 & 0.1122\\
    \midrule
    \multicolumn{9}{c}{$L=512,T=96$}\\
    \midrule
    \b{LSTM} & 0.7134 & 0.3811 & 0.2001  & \b{0.1843} & 0.2719 & 0.2809 & 0.3868 & 0.4069\\
    \b{LSTNet} & 0.8501 & 0.2931 & 0.1617 & \b{0.1595} & 0.2360 & 0.3235 & 0.3202 & 0.3397\\
    \b{Transformer} & 2.7635 & 3.0211 & 0.5371 & 1.3160 & 1.7465 & 1.1508 & \b{0.4568} & 0.6904\\
    \b{Autoformer} & 0.7147 & 0.4788 & 0.5125 & 0.4777 & 0.3339 & 0.3632 & 0.2721 & \b{0.2704}\\
    \b{Informer} & 3.2822 & 3.1977 & 2.9025 & 3.6369 & 1.2768 & 1.4782 & \b{1.1671} & 2.0538\\
    \b{TimesNet} & 0.0966  & 0.0931 & 0.0973 & 0.0914 & \b{0.0891} & 0.0985 & 0.1129 & 0.1203\\
    \b{PatchTST} & 0.0874 & \b{0.0874} & 0.0897 & 0.0876 & 0.0908 & 0.0951 & 0.1014 & 0.1053\\
    \bottomrule
    \end{tabular}
    }}
\end{table}

\begin{table}[H]
    \centering
    \caption{Learning rate tuning over the validation set on the dataset \ihom. The metric is NMSE, and the best learning rates are highlighted.}
    \label{tab:secondHyp}
    \resizebox{\textwidth}{!}{
    \edit{
    \begin{tabular}{ccccccccc}
    \toprule
    \b{Model} & $10^{-3}$ & $5\times 10^{-4}$ & $10^{-4}$ & $5\times 10^{-5}$ & $10^{-5}$ & $5\times10^{-6}$ & $10^{-6}$ & $5\times 10^{-7}$\\
    \midrule
    \multicolumn{9}{c}{$L=512,T=4$}\\
    \midrule
    \b{LSTM} & 0.2058 & 0.1832 & \b{0.1554} & 0.1618 & 0.2546 & 0.3488 & 0.5317 & 0.5421 \\
    \b{LSTNet} & 0.2145 & 0.1604 & \b{0.1490} & 0.1819 & 0.3139 & 0.3417 & 0.4325 & 0.4871\\
    \b{Transformer} & 0.6288 & 0.5218 & 0.2916 & \b{0.2189} & 0.2717 & 0.2736 & 0.3208 & 0.3363 \\
    \b{Autoformer} & 0.3000 & 0.2691 & \b{0.2050} & 0.2067 & 0.2350 & 0.2452 & 0.3032 & 0.3404\\ 
    \b{Informer} & 0.6004 & 0.5635 & 0.3656 & 0.2733 & \b{0.2078} & 0.2143 & 0.3040 & 0.3499\\
    \b{TimesNet} & 0.0835 & \b{0.0804} & 0.0933 & 0.0885 & 0.0976 & 0.1105 & 0.1863 & 0.2157\\
    \b{PatchTST} & 0.1049 & \b{0.0908} & 0.0987 & 0.1007 & 0.1099 & 0.1196 & 0.1466 & 0.1679\\
    \midrule
    \multicolumn{9}{c}{$L=512,T=48$}\\
    \midrule
    \b{LSTM} & 0.3416 & \b{0.3173} & 0.3278 & 0.3371 & 0.3558 & 0.3897 & 0.5408 & 0.5500\\
    \b{LSTNet} & 0.2479 & \b{0.2456} & 0.2631 & 0.2912 & 0.4025 & 0.4197 & 0.4526 & 0.4988\\
    \b{Transformer} & 0.6229 & 0.3294 & 0.2991 & 0.3052 & 0.2859 & \b{0.2834} & 0.3273 & 0.3404\\
    \b{Autoformer} & 0.2992 & 0.2806 & 0.2806 & \b{0.2762} & 0.3030 & 0.3182 & 0.3428 & 0.3643\\
    \b{Informer} & 0.6450 & 0.5477 & 0.4009 & 0.3403 & \b{0.2319} & 0.2385 & 0.3393 & 0.3826\\
    \b{TimesNet} & 0.1439 & 0.1327 & 0.1258 & \b{0.1234} & 0.1322 & 0.1432 & 0.2065 & 0.2307\\
    \b{PatchTST} & 0.1973 & \b{0.1889} & 0.1994 & 0.1927 & 0.1986 & 0.2017 & 0.2176 & 0.2283\\
    \midrule
    \multicolumn{9}{c}{$L=512,T=96$}\\
    \midrule
    \b{LSTM} & 0.2997 & 0.2389 & \b{0.2321} & 0.2402 & 0.3000 & 0.3673 & 0.5341 & 0.5444\\
    \b{LSTNet} & 0.2422 & 0.2350 & \b{0.2257} & 0.2566 & 0.3508 & 0.3741 & 0.4422 & 0.4911\\
    \b{Transformer} & 0.6371 & 0.3380 & 0.2984 & 0.3042 & 0.2915 & \b{0.2897} & 0.3288 & 0.3407\\
    \b{Autoformer} & 0.2990 & 0.2743 & \b{0.2401} & 0.2448 & 0.2766 & 0.2868 & 0.3327 & 0.3564\\
    \b{Informer} & 0.6068 & 0.5414 & 0.4018 & 0.3598 & \b{0.2571} & 0.2774 & 0.3366 & 0.3705\\
    \b{TimesNet} & 0.1503 & 0.1443 & 0.1325 & \b{0.1299} & 0.1523 & 0.1745 & 0.2247 & 0.2515\\
    \b{PatchTST} & 0.1855 & 0.1830 & 0.1856 & \b{0.1809} & 0.1883 & 0.1911 & 0.2037 & 0.2114\\
    \bottomrule 
    \end{tabular} 
    }}
\end{table}


The training is carried out using the above model-specific learning rates. As described in Section~\ref{sec:ed}, we use an early stopping criteria of no improvement in the validation set metrics (also called \emph{patience}) for 5 epochs. On the other hand, all training terminates in 20 epochs whether or not the patience criteria is met. The epoch at which each model's training terminates is given in Table~\ref{tab:epoch}.\par
\begin{table}[H]
    \centering
    \caption{Termination epoch for each trained model. The asterisk ($*$) indicates that early stopping was not triggered.}
    \label{tab:epoch}
    \edit{
    \begin{tabular}{lccc|ccc}
         \toprule
         \textbf{Model} & \multicolumn{3}{c}{\ihet} & \multicolumn{3}{c}{\ihom}\\
         & $T=4$ &  $T=48$ &  $T=96$ & $T=4$ &  $T=48$ &  $T=96$\\
         \midrule
         \b{LSTM} & 18 & $20^*$ & $20^*$ & 4 & 4 & 17\\
         \b{LSTNet} & 2 & 12 & 6 & 10 & 6 & 8\\
         \b{Transformer} & 4 & 4 & 4 & $20^*$ & $20^*$ & $20^*$\\
         \b{Autoformer} & $20^*$ & $20^*$ & $20^*$ & $20^*$ & 1 & 5\\
         \b{Informer} &2 & 1 &11 & 6 & 3 & 11\\
         \b{TimesNet} & 3 & 3 & 1 & $20^*$ & $7$ & 15\\
         \b{PatchTST} & 1 & 5 & 3 & 8 & 11 & 1\\
         \bottomrule
    \end{tabular}
    }
\end{table}

\begin{figure}[H]
    \centering
    \hfill
    \begin{subfigure}[b]{0.4\linewidth}
        \includegraphics[height=3.65cm, keepaspectratio]{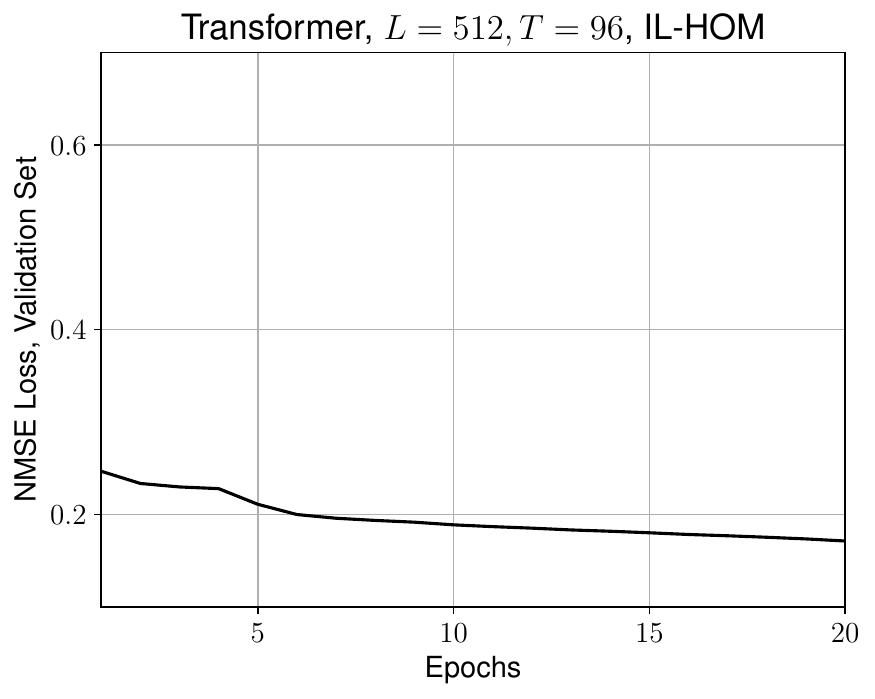}
        \caption{}
    \end{subfigure}
    \begin{subfigure}[b]{0.4\linewidth}
        \includegraphics[height=3.65cm, keepaspectratio]{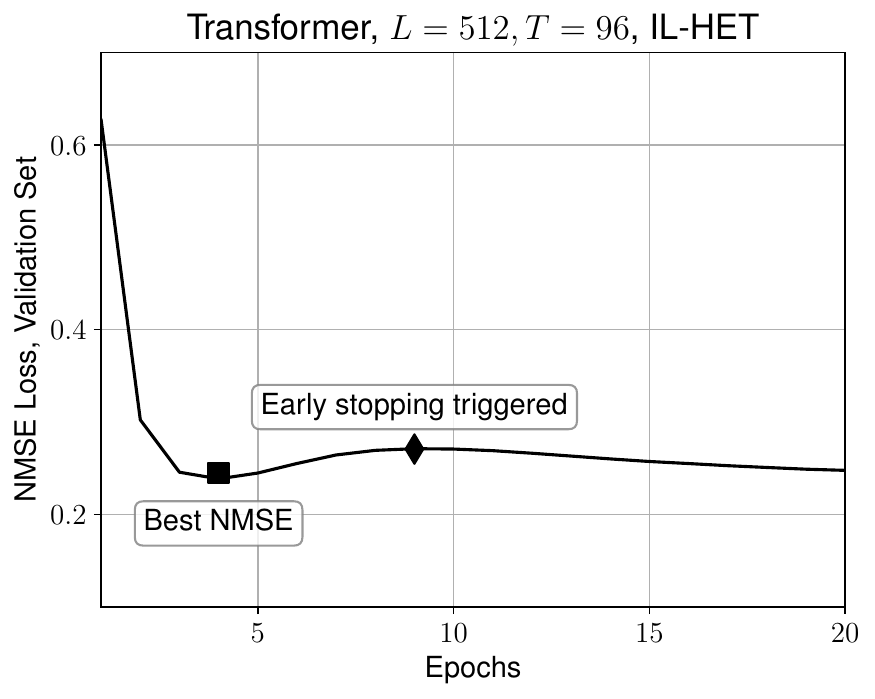}
        \caption{}
    \end{subfigure}
    \hfill
    \caption{Training instability in Transformer training on \ihet\ compared to \ihom.}
    \label{fig:instability}
\end{figure}

The observations of Table~\ref{tab:epoch} can be explained by the fact that training of \ihet\ is more unstable than \ihom, which leads to early stopping being triggered very early into the training run. This is further verified by letting Transformer for $L=512,T=96$ train to completion without early stopping on \ihet, which reveals instability in training \edit{after epoch 4} (see Figure~\ref{fig:instability}).

Similar to the base models, the learning rate for TimeFM was tuned over the validation set of \edit{all lookaheads $(T=4,48,96)$} with choices \edit{$\texttt{lr}=\{10^{-3},5\times10^{-4},10^{-4},5\times10^{-5},10^{-5},5\times10^{-6},10^{-6},5\times10^{-7}\}$} over 1 epoch. The results are reported in Tables~\ref{tab:firstHyp_timesfm} and~\ref{tab:secondHyp_timesfm}. \edit{With $T=(48,96)$, we perform additional experiments with $\texttt{lr}=\{10^{-7},5\times10^{-8}\}$}. \par

\begin{table}[H]
    \centering
    \caption{Learning rate tuning over the validation set of the problem for TimesFM. The metric is NMSE, and the best learning rates are highlighted.}
    \label{tab:firstHyp_timesfm}
    \edit{
    \resizebox{\textwidth}{!}{
    \begin{tabular}{ccccccccc}
    \toprule
    \textbf{Model} & $10^{-3}$ & $5\times 10^{-4}$ & $10^{-4}$ & $5\times 10^{-5}$ & $10^{-5}$ & $5\times10^{-6}$ & $10^{-6}$ & $5\times 10^{-7}$\\
    \midrule
    $(L,T)$ & \multicolumn{8}{c}{\ihom}\\
    \midrule
    \b{$(512,4)$} & 639.06 & 6.7198 & 0.0177 & 0.0166 & \b{0.0074} & 0.0076 & 0.0077 & 0.0082 \\
    \b{$(512,48)$} & 656.45 & 7.7152 & 0.0875 & 0.0512 & 0.0384 & \b{0.0365} & 0.0383 & 0.0404 \\
    \b{$(512,96)$} & 169.77 & 0.9800 & 0.1185 & 0.0483 & 0.0403 & \b{0.0401} & 0.0413 & 0.0432 \\
    \midrule
    $(L,T)$ & \multicolumn{8}{c}{\ihet}\\
    \midrule
    \b{$(512,4)$} & 2285.12 & 89.396 & 0.0331 & 0.0252 & 0.0144 & 0.0121 & \b{0.0118} & 0.0124 \\
    \b{$(512,48)$} & 247.92 & 10.034 & 0.0814 & 0.0724 & 0.0486 & 0.0473 & 0.0398 & \b{0.0369} \\
    \b{$(512,96)$} & 753.21 & 1.9964 & 0.0539 & 0.0568 & 0.0465 & 0.0462 & 0.0383 & \b{0.0369} \\
    \bottomrule
    \end{tabular}
    }
    }
\end{table}

\begin{table}[H]
    \centering
    \caption{Additional learning rate tuning over the validation set of the problem for TimesFM for \ihet\ with $(512,48)$ and $(512,96)$. The metric is NMSE, and the best learning rates are highlighted.}
    \label{tab:secondHyp_timesfm}
    \edit{
    \begin{tabular}{cccccccc}
    \toprule
    \textbf{Model} & $10^{-7}$ & $5\times 10^{-8}$ \\
    \midrule
    $(L,T)$ & \multicolumn{2}{c}{\ihet}\\
    \midrule
    \b{$(512,48)$} & 0.04121 & 0.04711 \\
    \b{$(512,96)$} & 0.04027 & 0.04585 \\
    \bottomrule
    \end{tabular}
    }
\end{table}

Early-stopping based on \textit{patience} is also implemented in finetuning the TimesFM model where validation is performed to compute the progress of the finetuning in terms of the NMSE loss every 1,000 steps in each epoch. \textit{Patience} is increased by one if the NMSE loss does not decrease and the so-called early termination of the current epoch is executed if \textit{patience} reaches five. In our finetuning experience, the early termination condition was never triggered. 

\section{Predictions}
\label{sec:reconst}
\begin{figure}[H]
    \centering
    \begin{tabular}{cccc}
        \begin{subfigure}{0.28\textwidth}
            \includegraphics[width=\linewidth]{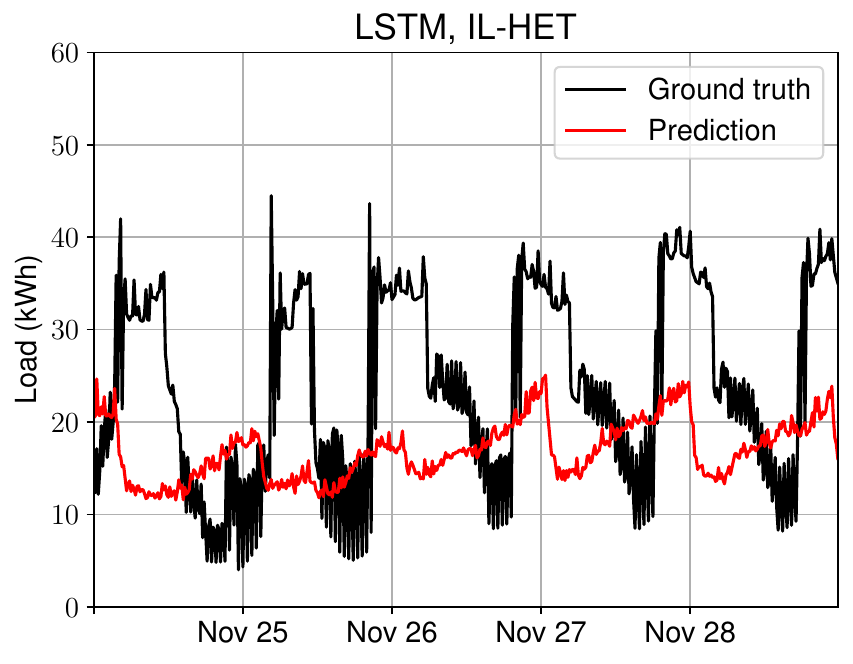}
            \caption{}
        \end{subfigure} &
        \begin{subfigure}{0.28\textwidth}
            \includegraphics[width=\linewidth]{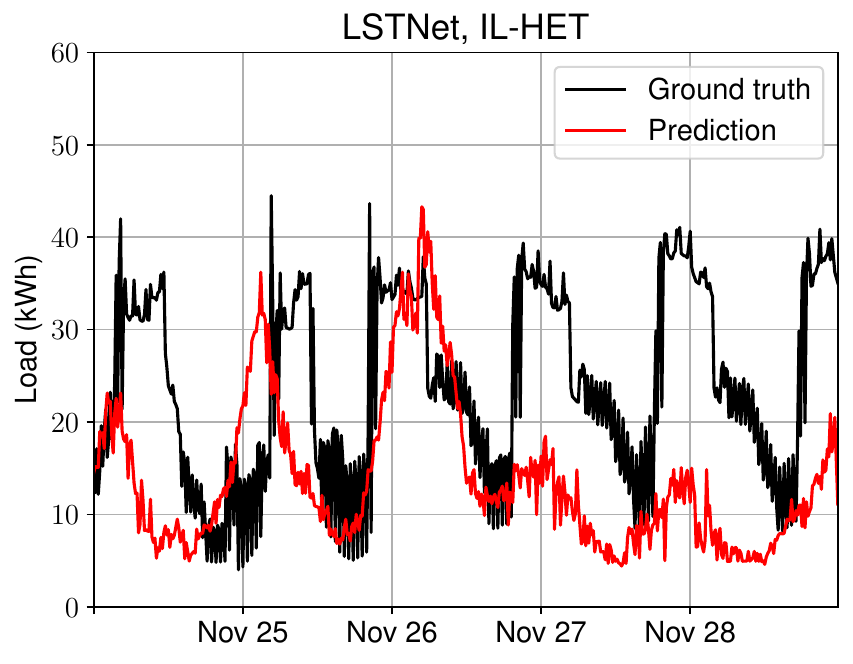}
            \caption{}
        \end{subfigure} &
        \begin{subfigure}{0.28\textwidth}
            \includegraphics[width=\linewidth]{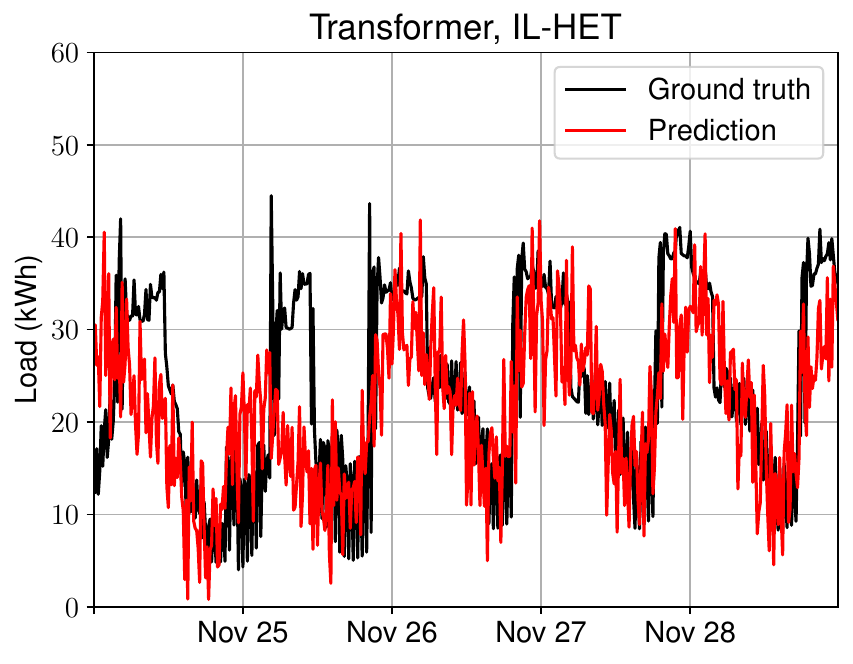}
            \caption{}
        \end{subfigure}
        \\
        \begin{subfigure}{0.28\textwidth}
            \includegraphics[width=\linewidth]{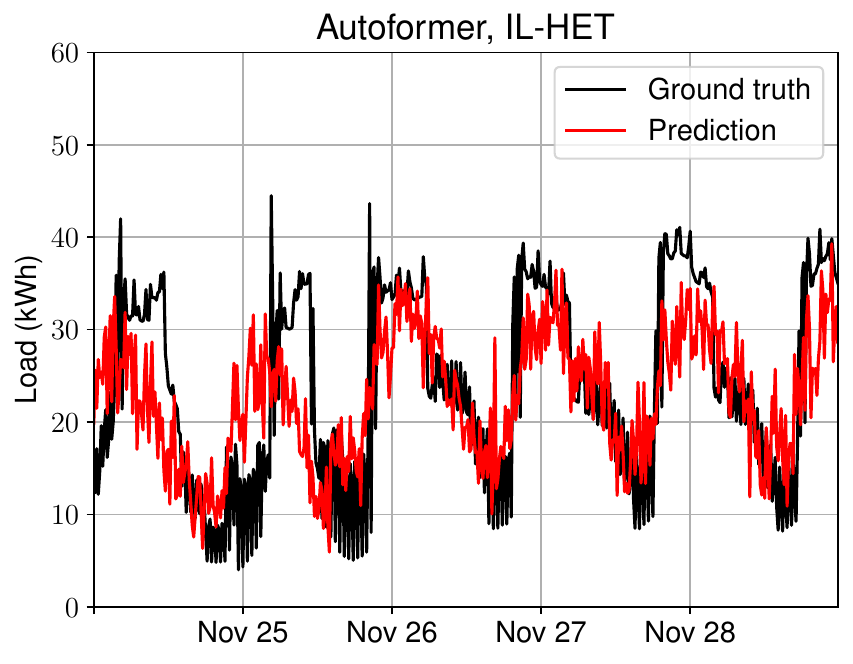}
            \caption{}
        \end{subfigure} &
        \begin{subfigure}{0.28\textwidth}
            \includegraphics[width=\linewidth]{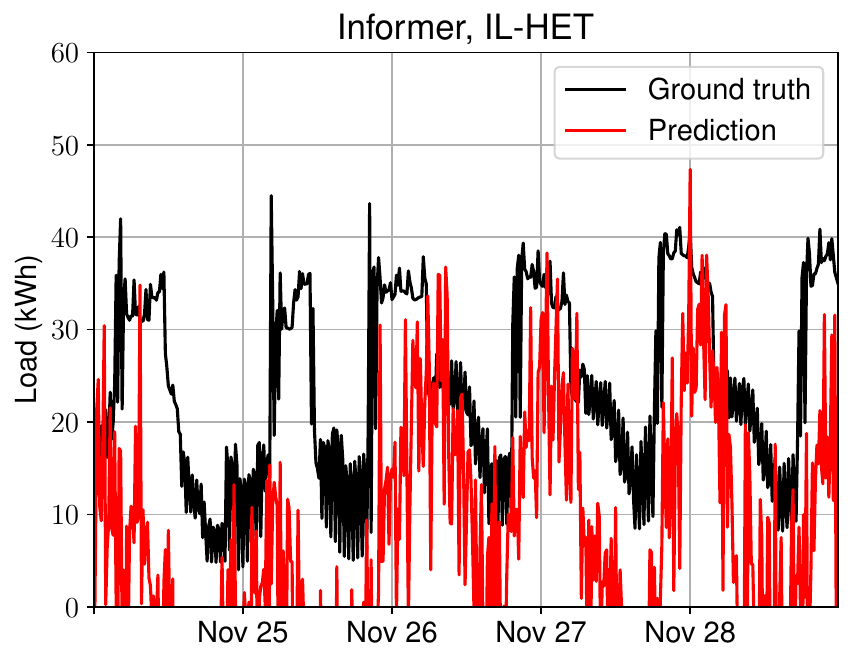}
            \caption{}
        \end{subfigure} &
        \begin{subfigure}{0.28\textwidth}
            \includegraphics[width=\linewidth]{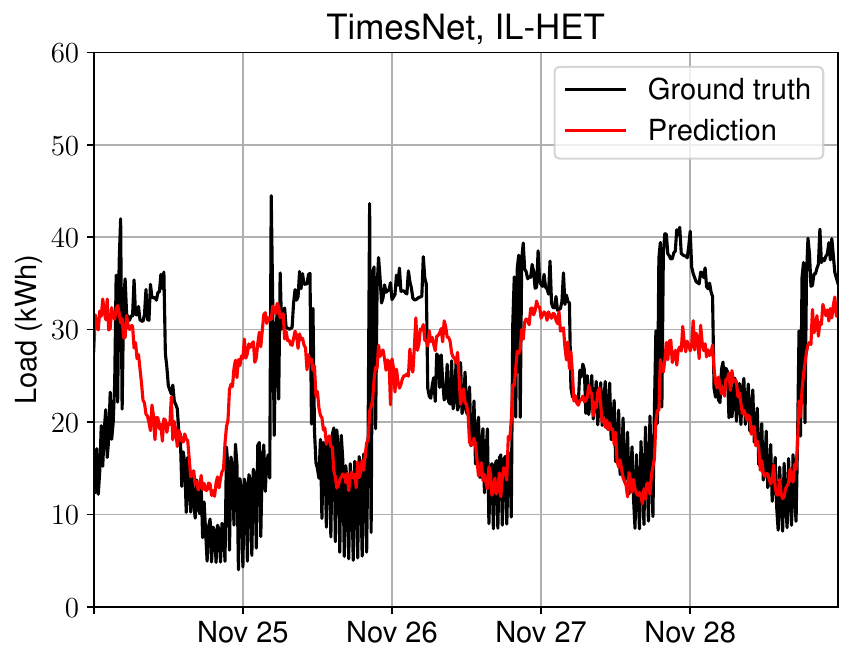}
            \caption{}
        \end{subfigure} \\
        \begin{subfigure}{0.28\textwidth}
            \includegraphics[width=\linewidth]{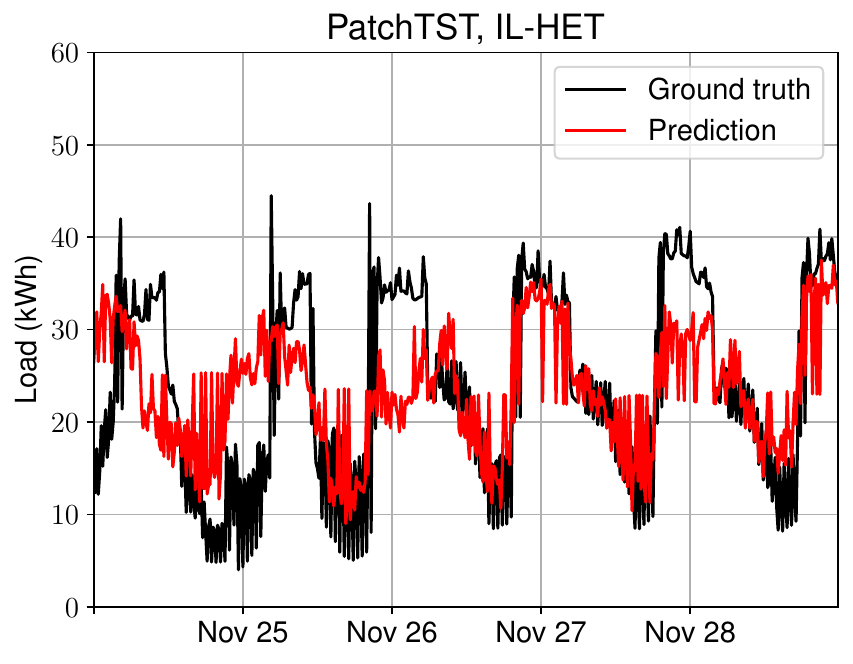}
            \caption{}
        \end{subfigure} & 
        \begin{subfigure}{0.28\textwidth}
            \includegraphics[width=\linewidth]{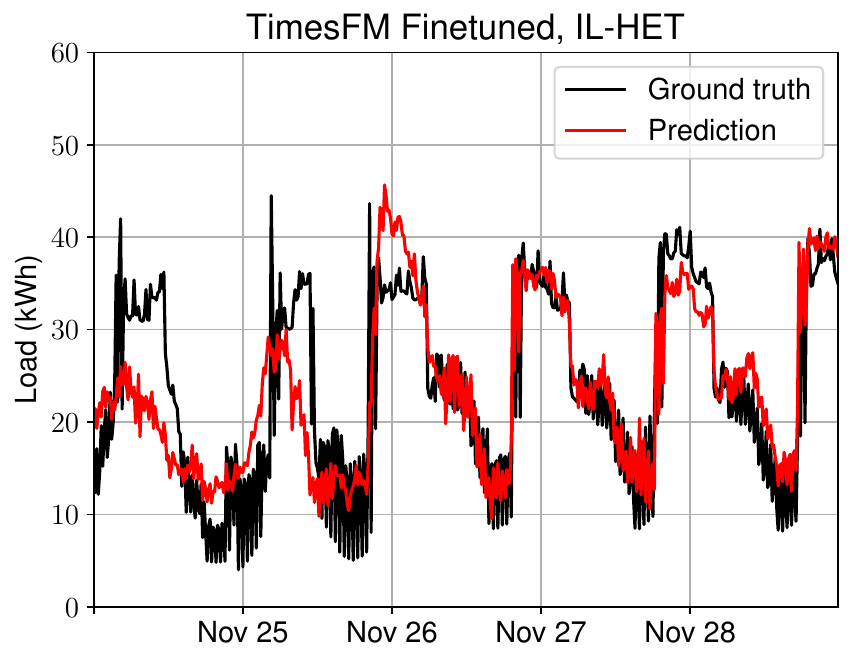}
            \caption{}
        \end{subfigure} &
        \begin{subfigure}{0.28\textwidth}
            \includegraphics[width=\linewidth]{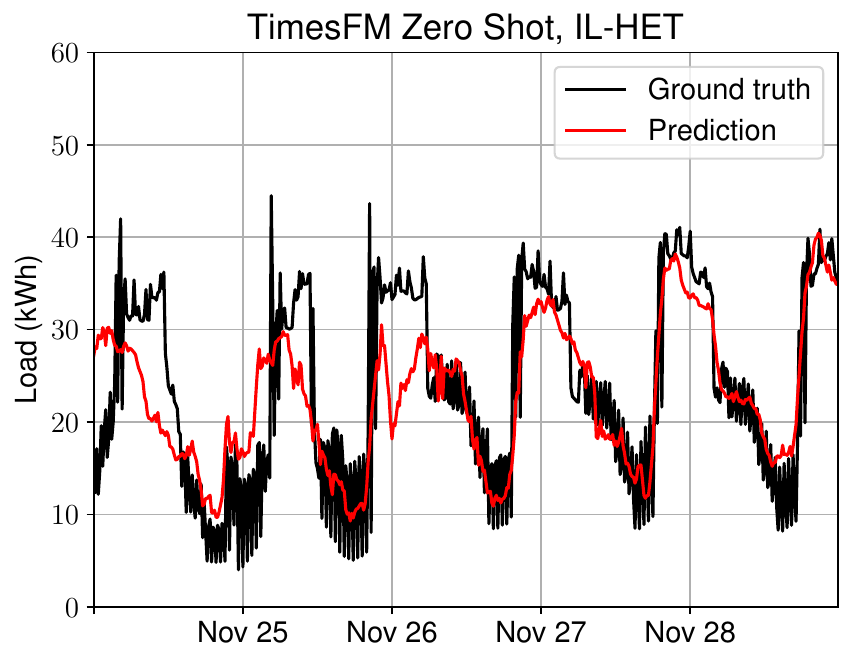}
            \caption{}
        \end{subfigure} \\
        \begin{subfigure}{0.28\textwidth}
            \includegraphics[width=\linewidth]{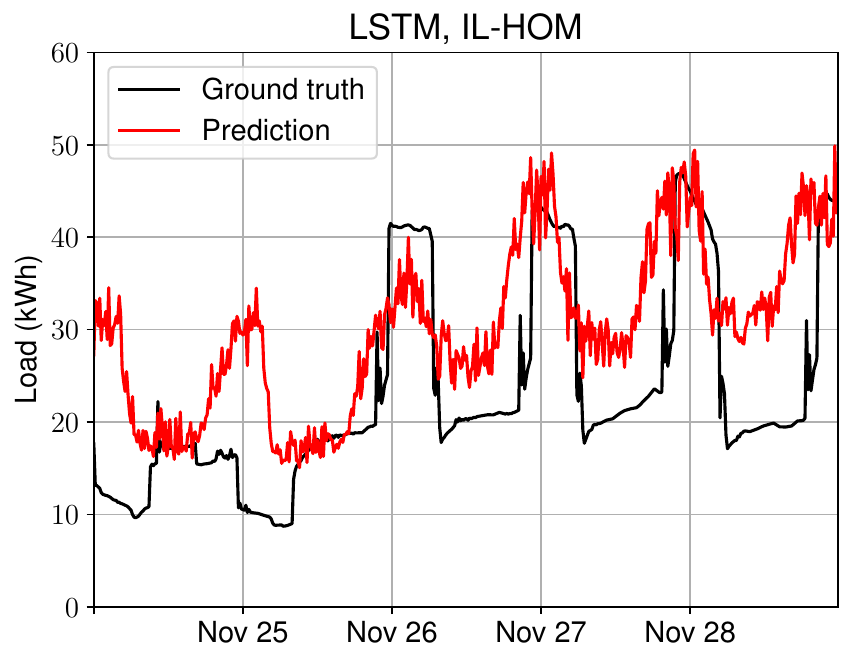}
            \caption{}
        \end{subfigure} &
        \begin{subfigure}{0.28\textwidth}
            \includegraphics[width=\linewidth]{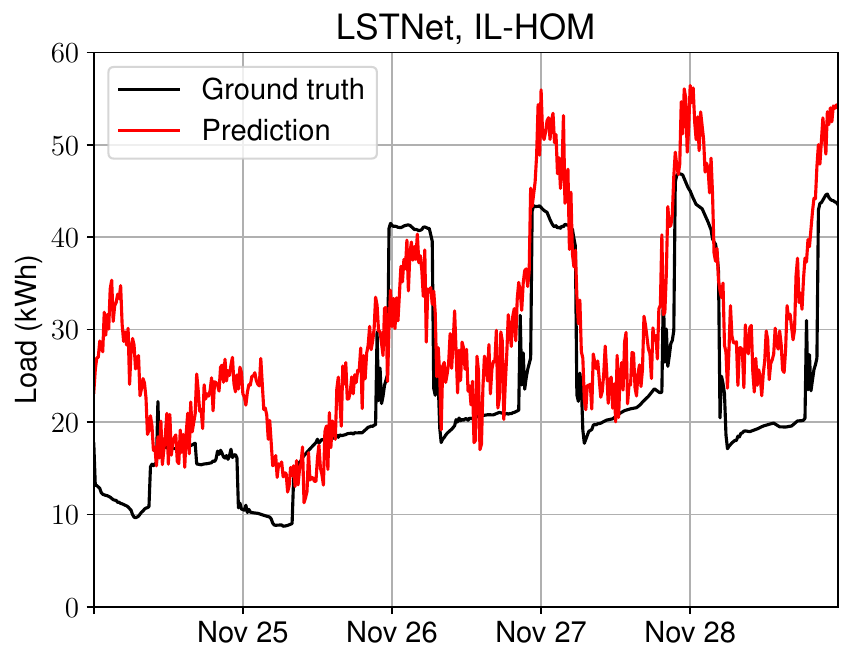}
            \caption{}
        \end{subfigure} &
        \begin{subfigure}{0.28\textwidth}
            \includegraphics[width=\linewidth]{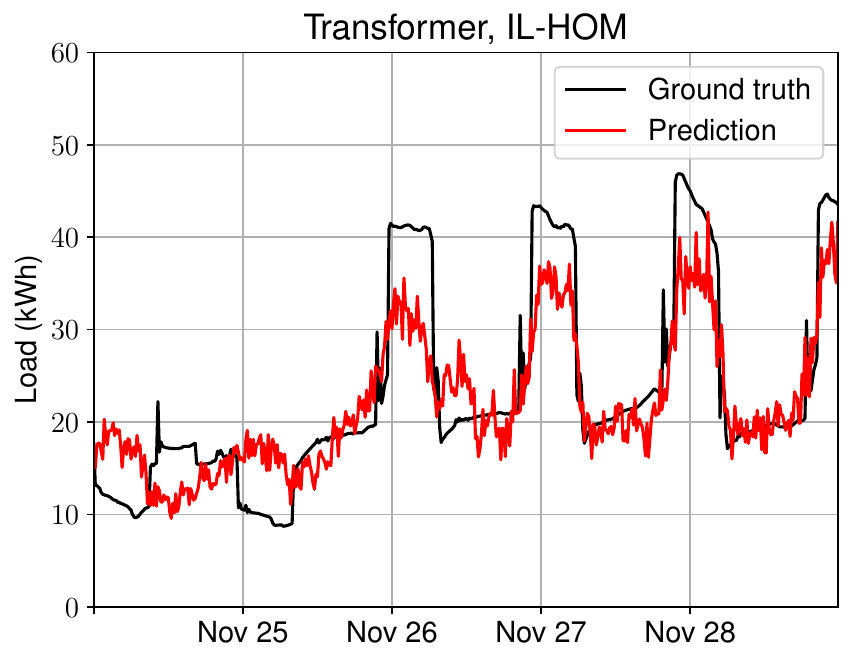}
            \caption{}
        \end{subfigure}\\
        \begin{subfigure}{0.28\textwidth}
            \includegraphics[width=\linewidth]{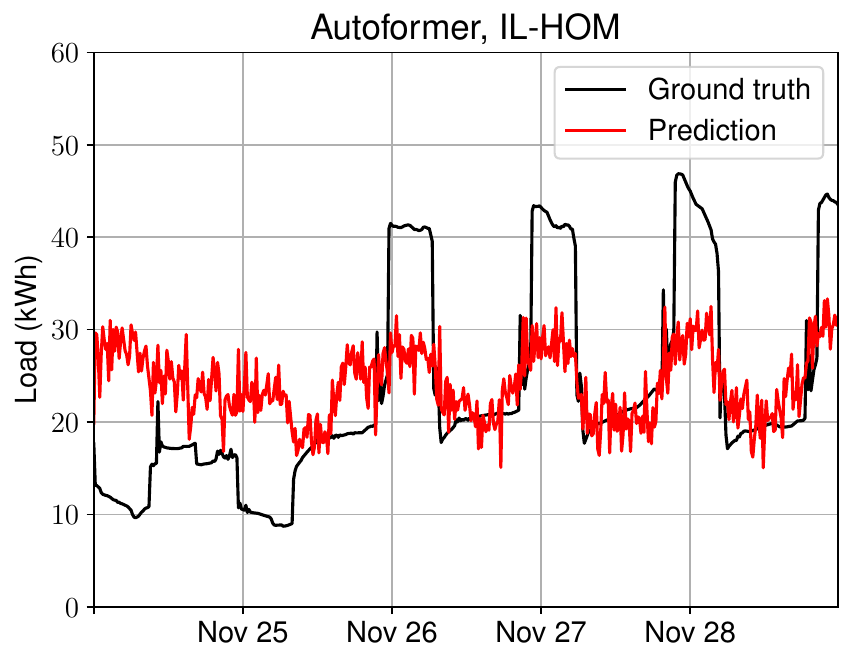}
            \caption{}
        \end{subfigure} &
        \begin{subfigure}{0.28\textwidth}
            \includegraphics[width=\linewidth]{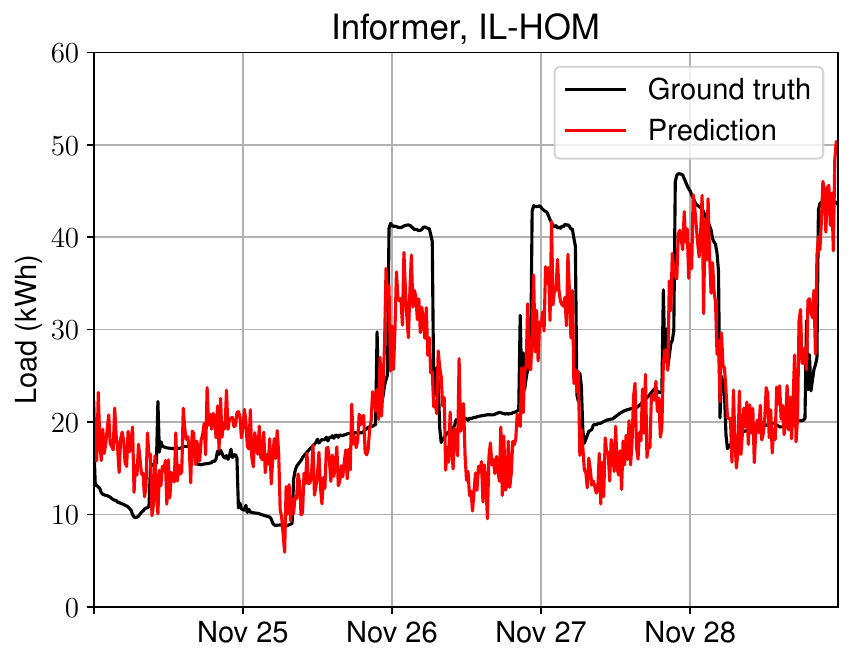}
            \caption{}
        \end{subfigure} &
        \begin{subfigure}{0.28\textwidth}
            \includegraphics[width=\linewidth]{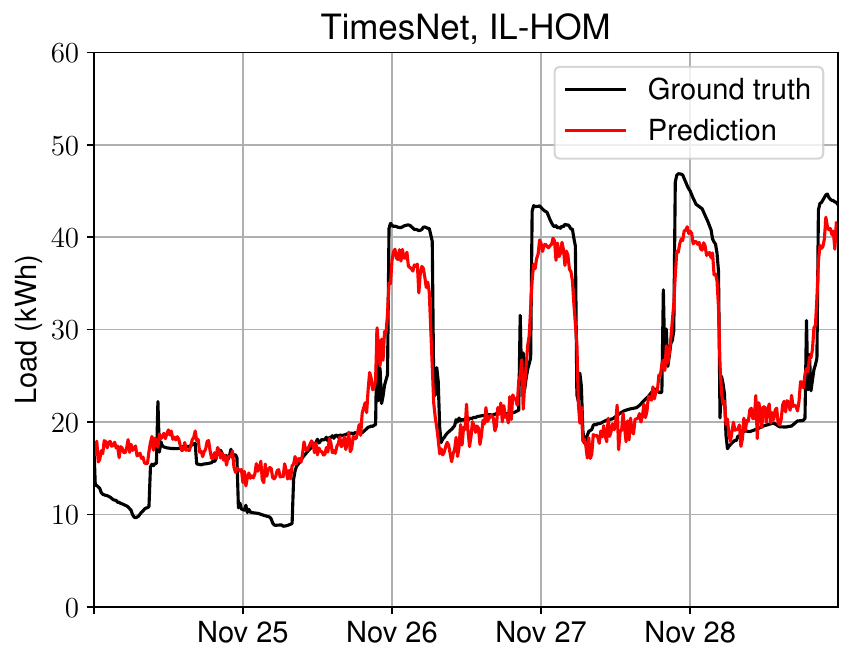}
            \caption{}
        \end{subfigure} \\
        \begin{subfigure}{0.28\textwidth}
            \includegraphics[width=\linewidth]{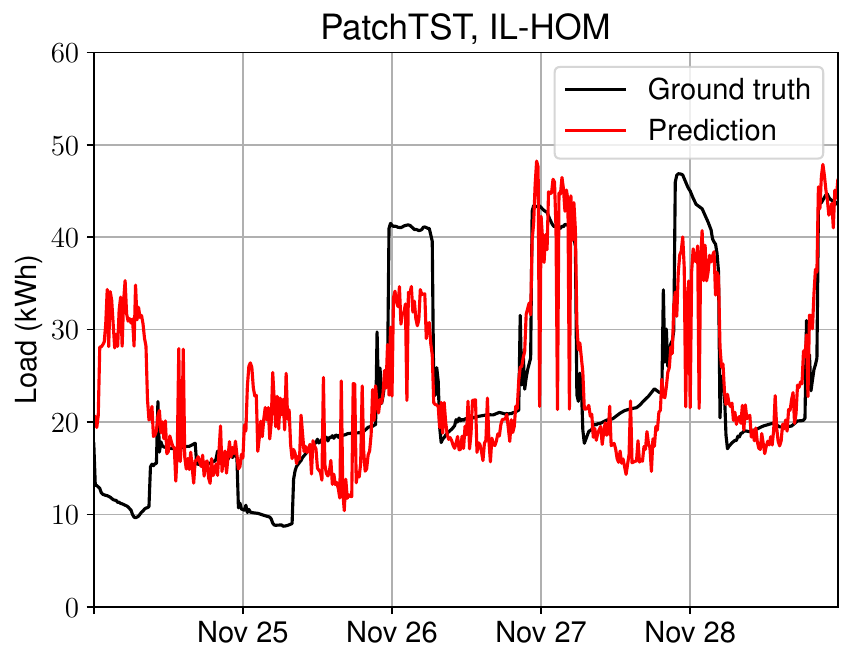}
            \caption{}
        \end{subfigure} &
        \begin{subfigure}{0.28\textwidth}
            \includegraphics[width=\linewidth]{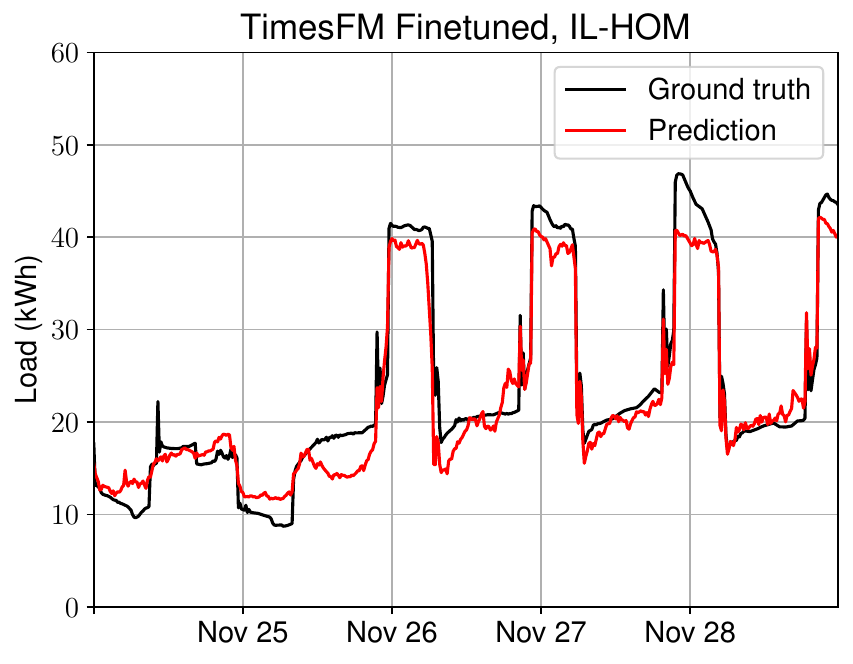}
            \caption{}
        \end{subfigure} &
        \begin{subfigure}{0.28\textwidth}
            \includegraphics[width=\linewidth]{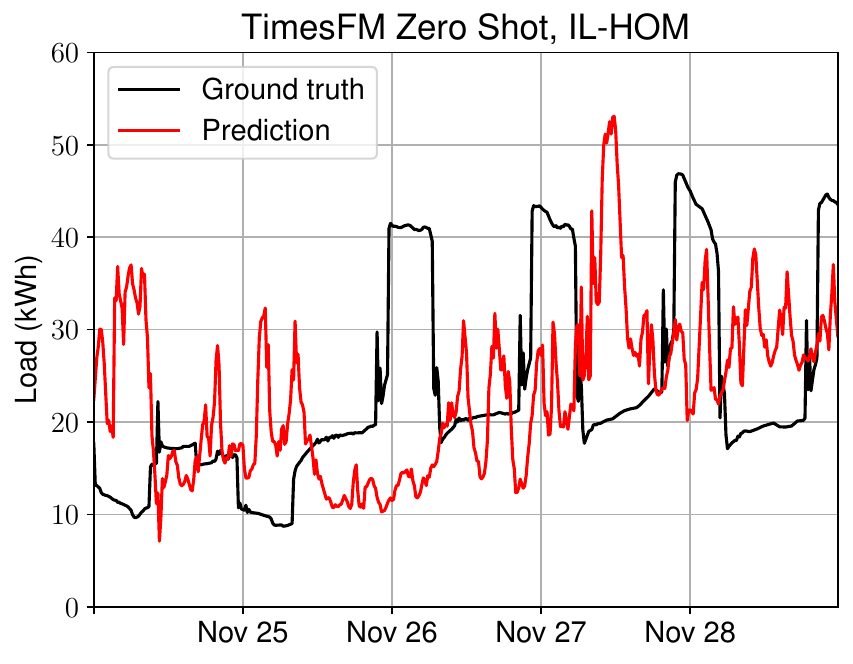}
            \caption{}
        \end{subfigure} \\
    \end{tabular}
    \caption{Predictions of the test set loads for two selected buildings from \ihom\ and \ihet\ datasets for $T=512, L=48$, respectively, for the best models reported in Table~\ref{table:features}.}
    \label{fig:reconst}
\end{figure}

In Figure~\ref{fig:reconst}, we show the predictions for the first 500 points of the $5^{\text{th}}$ customer in both the datasets. These visualizations offer insights into how well the trained models capture the underlying patterns in individual building data. However, it is important to note that the predictions for a single building may not fully represent the model's performance across the entire test set.

\end{document}